\documentclass{article}
\usepackage{spconf}
\usepackage{amsfonts}
\usepackage{amssymb}
\usepackage{bm}
\usepackage{enumitem}
\usepackage{float}
\usepackage{url}
\usepackage{multirow}
\usepackage{subcaption}
\usepackage{cite}
\usepackage{amsmath,amssymb,amsfonts}
\usepackage{graphicx}
\usepackage{textcomp}
\usepackage{xcolor}
\usepackage{booktabs}
\usepackage{float}
\usepackage{adjustbox}
\usepackage{hyperref}
\usepackage{comment}
\usepackage{algorithm}
\usepackage{algpseudocode} 
\usepackage[table]{xcolor}
\usepackage[font=footnotesize,skip=2pt]{caption}
\usepackage{bbm}

\makeatletter
\newcommand{\StatePar}[1]{%
  \State\parbox[t]{\dimexpr\linewidth-\ALG@thistlm}{\strut #1\strut}%
}
\makeatother

\makeatletter
\renewcommand{\paragraph}{\@startsection{paragraph}{4}{\z@}%
  {0.75ex \@plus 0.5ex \@minus 0.2ex}
  {-1em}
  {\normalfont\normalsize\bfseries}}
\makeatother
\makeatletter
\renewcommand{\ALG@beginalgorithmic}{\small} 
\makeatother

\DeclareMathOperator*{\argmax}{arg\,max}

\algrenewcommand\algorithmicrequire{\textbf{Input:}}
\algrenewcommand\algorithmicensure{\textbf{Output:}}
\algrenewcommand\algorithmiccomment[1]{\hfill\(\triangleright\) #1}
\makeatletter
\g@addto@macro\normalsize{%
  \setlength\abovedisplayskip{3pt}%
  \setlength\belowdisplayskip{3pt}%
  \setlength\abovedisplayshortskip{1pt}%
  \setlength\belowdisplayshortskip{1pt}%
}
\makeatother

\makeatletter
\renewcommand{\section}{\@startsection{section}{1}{\z@}%
  {1.5ex}
  {1ex}
  {\normalfont\normalsize\bfseries}} 
\makeatother


\title{ARMOR: Agentic Reasoning for Methods Orchestration and Reparameterization for Robust Adversarial Attacks}
\name{
\begin{tabular}{c}
Gabriel Jun Rong Lee$^{1}$ \qquad Christos Korgialas$^{2}$ \qquad Dion Jia Xu Ho$^{3}$ \\
Pai Chet Ng$^{1}$ \qquad Xiaoxiao Miao$^{4}$ \qquad Konstantinos N. Plataniotis$^{5}$
\end{tabular}
}
\address{
$^{1}$Infocomm Technology Cluster, Singapore Institute of Technology, Singapore \\
$^{2}$Dept. of Informatics, Aristotle University of Thessaloniki, Greece \\
$^{3}$Dept. of Applied Physics and Applied Mathematics, Columbia University, USA \\
$^{4}$Div. of Natural and Applied Sciences, Duke Kunshan University, China \\
$^{5}$Dept. of Electrical and Computer Engineering, University of Toronto, Canada
}

\begin{document}
\ninept

\maketitle

\begin{abstract} 
Existing automated attack suites operate as static ensembles with fixed sequences, lacking strategic adaptation and semantic awareness. This paper introduces the Agentic Reasoning for Methods Orchestration and Reparameterization (ARMOR) framework to address these limitations. ARMOR orchestrates three canonical adversarial primitives, Carlini-Wagner (CW), Jacobian-based Saliency Map Attack (JSMA), and Spatially Transformed Attacks (STA) via Vision Language Models (VLM)-guided agents that collaboratively generate and synthesize perturbations through a shared ``Mixing Desk". Large Language Models (LLMs) adaptively tune and reparameterize parallel attack agents in a real-time, closed-loop system that exploits image-specific semantic vulnerabilities. On standard benchmarks, ARMOR achieves improved cross-architecture transfer and reliably fools both settings, delivering a blended output for blind targets and selecting the best attack or blended attacks for white-box targets using a confidence-and-SSIM score. 
\end{abstract}


\section{Introduction} \label{sec:intro}

The proliferation of high-fidelity synthetic media, or deepfakes, poses a significant challenge to information integrity and digital security \cite{tolosana2020deepfakes, mirsky2021creation}. To counter this, advanced deepfake detection systems \cite{ng2024can} based on CNNs (like ResNet-50 \cite{he2016deep}, DenseNet-121 \cite{huang2017densely}) and Vision Transformers (ViT-B/16) \cite{dosovitskiy2020image} have been developed, showing high accuracy on benchmarks \cite{rossler2019faceforensicspp}. However, their reliability in adversarial settings is questionable, as DNNs are susceptible to adversarial examples modified with small, often imperceptible perturbations designed to induce misclassification \cite{szegedy2014intriguing, goodfellow2015advexamples}. This vulnerability threatens the practical deployment of deepfake detectors.

Prior work on generating adversarial attacks has focused on single-agent optimization algorithms, all of which use gradient information. This includes hard-constraint attacks like the Fast Gradient Sign Method (FGSM) \cite{goodfellow2015advexamples} and Projected Gradient Descent (PGD) \cite{madry2018towards}, which are explicitly designed to satisfy a predefined $\epsilon$ attack budget. Other gradient-based methods, such as the CW attack \cite{carlini2017towards}, the JSMA \cite{papernot2016limitations}, and STA examples \cite{xiao2018spatially}, aim to minimize distortion or strategically modify minimal features. Despite their individual merits, these methods operate as static, non-adaptive processes and often exhibit poor transferability across different model architectures \cite{liu2017delving, papernot2017practical}. A comprehensive overview of these methods is available in recent surveys \cite{akhtar2018threat, serban2020adversarial}.


Recent breakthroughs in generative AI offer new paradigms for complex problem-solving. VLMs like Qwen2-VL \cite{bai2024qwenvl} enable high-level semantic reasoning by grounding textual concepts in visual data. Furthermore, agentic AI, where LLMs act as reasoning engines for autonomous agents \cite{xi2025rise, tran2025macollab}, has gained traction, allowing complex tasks to be solved by collaborative agent societies \cite{wu2024autogen} in areas like software development and science \cite{park2023generative}, and, relevantly, for hyperparameter optimization \cite{pmlr-v280-liu25c}. However, the application of these multi-agent systems to the strategic domain of adversarial machine learning remains largely unexplored. While some LLM-based automation exists for red-teaming \cite{deng2023llms}, it lacks real-time, dynamic collaboration between multiple attack algorithms.

\textbf{Motivation:} While automated attack suites like AutoAttack \cite{croce2020reliable} provide robustness benchmarks, these static ensembles lack real-time, semantic-aware adaptation, relying on fixed, well-tuned sequences, and each attack is, for the images it is assigned, no more effective than when run alone. 
\textbf{Why VLMs and LLMs?} VLMs extract image-grounded semantic cues (e.g., face regions and hair/skin textures) that inform where and how to allocate perturbations for high impact with minimal perceptual change. LLMs use iterative feedback (surrogate stability, SSIM) to adapt attack hyperparameters and perturbation mixing across iterations, enabling closed-loop strategy updates when optimization stagnates.
\textbf{Proposed vs. Prior work:} We introduce the Agentic Reasoning for Methods Orchestration and Reparameterization (ARMOR) framework, reframing attack generation as dynamic, collaborative, and semantic-aware optimization. ARMOR adaptively orchestrates CW \cite{carlini2017towards}, JSMA \cite{papernot2016limitations}, and STA \cite{xiao2018spatially} via a VLM-powered Analysis agent (Qwen2.5-VL-32B-Instruct-AWQ) to counter semantic blindness. Method agents run in parallel and are reparameterized by Advisor LLMs (Qwen3-32B-AWQ) to avoid strategic stagnation, and Critique agents in a closed-loop ``Mixing Desk.” This yields goal-directed coordination that static ensembles cannot achieve, extending ensemble ideas \cite{dong2018boosting} with intelligent strategy control. \textbf{Contributions:} We propose ARMOR, a VLM/LLM-driven multi-agent framework that orchestrates complementary perturbation geometries CW (dense), JSMA (sparse), and STA (geometric) covering modes that transfer differently across architectures~\cite{wang2024survey}. ARMOR introduces a continuous closed-loop that evaluates mixed candidates using black-box confidence, surrogate stability, and SSIM, and escalates strategy under stagnation. We show improved transfer-based black-box evasion of a blind ViT detector under a common $\ell_\infty$ budget and query limit.

\begin{algorithm}[t]
\caption{The ARMOR Iteration Loop}
\label{alg:armor_framework}
\begin{algorithmic}[1]
\State \textbf{Input:} $\bm{x}$, $p_{\text{bb}}$, $\mathcal{P}_{\text{surr}}$, $t_0$
\Statex \Comment{\textbf{Phase 1: Objective Formulation}}
\State \textbf{Initialize:}
\State \quad $\mathcal{R} \gets \textit{InfoAgent}(\text{VLM}(\bm{x}))$ \Comment{VLM analyzes image}
\State \quad $\bm{p}_{\text{base}} \gets \text{Evaluate}(\bm{x}, \mathcal{P}_{\text{surr}}, p_{\text{bb}})$ \Comment{Baseline predictions}
\State \quad $(\epsilon, \tau, t) \gets \textit{ConductorAgent}(\mathcal{R}, \bm{p}_{\text{base}}, t_0)$ \Comment{Set constraints}
\State \quad History $\mathcal{H} \leftarrow \emptyset$, Iteration $k \leftarrow 0$
\While{not converged and budget not exhausted}
\State $k \leftarrow k+1$
\Statex \Comment{\textbf{Phase 2: Parallel Perturbation Generation}}
\State $\bm{\Delta} \leftarrow \emptyset$
\For{method $m \in \{\text{CW, JSMA, STA}\}$}
  \State $\bm{\phi}_m \gets \textit{AdvisorAgent}(\mathcal{H}, m)$
  \State $\bm{\delta}_m \gets \text{Generate}_m(\bm{x}, t, \epsilon, \bm{\phi}_m, \mathcal{P}_{\text{surr}})$
  \State $\bm{\Delta} \leftarrow \bm{\Delta} \cup \{\bm{\delta}_m\}$
\EndFor

\Statex \Comment{\textbf{Phase 4: Adaptive Perturbation Ensemble}}
\State $\bm{w}^* \leftarrow \text{Optimize \eqref{eq:score} via randomized hill climbing}$
\State $\bm{x}_{\text{master}}^{(k)} \gets \bm{x}_{\text{master}}(\bm{w}^*)$ \Comment{Construct via \eqref{eq:mix}}

\Statex \Comment{\textbf{Phase 3: Critique \& Strategic Adaptation}}
\State $\bm{c}^{(k)} \leftarrow \textit{CritiqueAgents}(\bm{x}_{\text{master}}^{(k)})$ \Comment{bb success, surrogate transfer, SSIM}
\State $\mathcal{H} \leftarrow \mathcal{H} \cup \{(k, \bm{\Delta}, \bm{w}^*, \bm{c}^{(k)})\}$
\State $I_k \gets \{k\!-\!W\!+\!1,\ldots,k\}\cap\mathbb{N}^+$;\quad $p_{\text{bb}}^{(i)} \gets p_{\text{bb}}(t\,|\,\bm{x}_{\text{master}}^{(i)})$
\If{\textit{StrategistAgent} detects stagnation in $I_k$}
  \State $(\tau, \epsilon) \leftarrow \text{Escalate}(\tau, \epsilon)$
\EndIf
\EndWhile
\State \textbf{Output:} Blind: Return blended $\bm{x}_{\text{master}}^{(k)}$; White-box: Return the attack or blend with the highest confidence-and-SSIM score.

\end{algorithmic}
\end{algorithm}

\section{Preliminaries}\label{sec:preliminaries}

Let $\bm{x} \in \mathcal{X} \subseteq [0,1]^d$ be the original input image, and $t$ be the target class label. Let $f:\mathcal{X}\!\to\!\{1,\dots,K\}$ be a classifier on $\mathcal{X}$ with logits $Z(\bm{x})\in\mathbb{R}^K$. Our experiments are binary ($K{=}2$, \emph{real} or \emph{fake}). We denote the cross-entropy loss for target $t$ as $\mathcal{L}_{\mathrm{CE}}(\cdot,t)$. The projection operators are defined as:
\begin{equation}
\Pi_{[0,1]}(\bm y)=\operatorname{clip}(\bm y,0,1),\qquad
\Pi_{\infty,\epsilon}(\bm\delta)=\operatorname{clip}(\bm\delta,-\epsilon,\epsilon).
\label{eq:projections}
\end{equation}

\textbf{CW:} The original CW attack \cite{carlini2017towards}, which uses an input box constraint but no inherent $\ell_\infty$ budget, solves the $\ell_2$-regularized minimization problem:
\begin{equation}
\min_{\bm{\delta}}\ \|\bm{\delta}\|_2^2 + c\,\mathcal{L}_{\mathrm{adv}}(Z(\bm{x}+\bm{\delta}),t)
\quad \text{s.t.}\quad \bm{x}+\bm{\delta}\in[0,1]^d.
\label{eq:cw}
\end{equation}
We set $\mathcal{L}_{\mathrm{adv}}=\mathcal{L}_{\mathrm{CE}}$ and define $J(\bm{\delta})=\|\bm{\delta}\|_2^2 + c\,\mathcal{L}_{\mathrm{adv}}$. We enforce a $\ell_\infty$ budget by projecting updates, i.e. $\bm{\delta} \leftarrow \Pi_{\infty,\epsilon}\!\big(\bm{\delta}-\eta\nabla_{\bm{\delta}} J\big)$.

\textbf{JSMA:} A targeted, saliency-based method that perturbs a small set of features by selecting coordinates $(p^\star,q^\star)$ that maximize movement toward target logit $t$:
\begin{equation}
(p^\star,q^\star)=\argmax_{(p,q)}\ -\alpha_{pq}\,\beta_{pq}\ \ \text{s.t.}\ \ \alpha_{pq}>0,\ \beta_{pq}<0,
\label{eq:jsma}
\end{equation}
with $\alpha_{pq}=\sum_{i\in\{p,q\}}\frac{\partial Z_t}{\partial x_i}$ and $\beta_{pq}=\sum_{i\in\{p,q\}}\sum_{j\ne t}\frac{\partial Z_j}{\partial x_i}$. Every iteration, JSMA perturbs one or two selected pixels by a fixed, user-specified step with perturbations projected via $\Pi_{\infty,\epsilon}$ and $\Pi_{[0,1]}$.

\textbf{STA:} Constrains perturbations by optimizing a spatial flow field $f$ rather than enforcing an $\ell_p$-norm budget. The objective balances the adversarial loss against a flow smoothness prior, $\mathcal{L}_{\mathrm{flow}}$:
\begin{equation}
\min_{f}\quad \mathcal{L}_{\mathrm{adv}}\big(x_{\mathrm{adv}}(f), t\big) + \theta\,\mathcal{L}_{\mathrm{flow}}(f),
\label{eq:sta}
\end{equation}
where image $x_{\mathrm{adv}}(f) = \mathcal{W}(\bm{x}; f)$ is a differentiable warp of the input image $\bm{x}$. The hyperparameter $\theta$ controls the trade-off, with the smoothness term implicitly constraining visual distortion.

\section{Proposed Methodology}\label{sec:methodology}

We address the transferability gap of adversarial examples from ResNet-50 and DenseNet-121 to a black-box ViT-B/16 by framing attack generation as an agentic optimization process. Our framework, ARMOR, employs specialized AI agents to plan, generate, critique, and adapt perturbations in a closed loop (see Algorithm~\ref{alg:armor_framework}).

We operate in a transfer-based black-box setting, where the attacker has no access to the target ViT-B/16. Instead, the attack is guided by gradients from a set of white-box surrogate models, $\mathcal{P}_{\text{surr}}$, consisting of a ResNet-50 and a DenseNet-121. To optimize for transferability, we use the mean target class probability across this surrogate ensemble as a differentiable proxy objective:
\begin{equation}
\bar{p}_{\text{surr}}(t | \bm{x}) = \frac{1}{|\mathcal{P}_{\text{surr}}|} \sum_{f \in \mathcal{P}_{\text{surr}}} p_f(t | \bm{x}).
\label{eq:surr_mean}
\end{equation}

\subsection{Phase 1: Agentic Reconnaissance \& Objective Formulation}

The attack commences with agent-driven reconnaissance. The \textit{InfoAgent} (VLM - Qwen2.5-VL-32B-Instruct-AWQ) analyzes the source image $\bm{x}$ to produce a semantic report $\mathcal{R}$. The \textit{ConductorAgent} (LLM - Qwen3-32B-AWQ) then synthesizes this report with predictions $\bm{p}_{\text{base}}$ from the baseline model to define the global attack objective as a constrained optimization problem, dynamically setting the key trade-off parameters $(\epsilon, \tau)$:
\begin{equation}
\max_{\bm{\delta}}\ p_{\text{bb}}(t\,|\, \bm{x}+\bm{\delta})
\quad
\text{s.t.} \quad \|\bm{\delta}\|_{\infty}\le \epsilon \ \text{and} \
\mathrm{SSIM}(\bm{x},\bm{x}+\bm{\delta})\ge \tau,
\label{eq:bbgoal}
\end{equation}
 where $\mathrm{SSIM}:\mathcal{X}\times\mathcal{X}\to[0,1]$ denotes the Structural Similarity Index, with $\mathrm{SSIM}=1$ for identical images. By tuning the maximum $\ell_{\infty}$-norm, $\epsilon\in(0,1]$, and minimum SSIM threshold $\tau\in[0,1]$, ARMOR balances adversarial potency against perceptual fidelity tailored to the image content.

\subsection{Phase 2: Parallel Perturbation Generation}

Each \textit{MethodAgent}, a specialized CW, JSMA, or STA attack (Section~\ref{sec:preliminaries}), generates a single candidate perturbation, $\bm{\delta}_m$, for subsequent mixing. Agents operate in parallel, receiving $(\bm{x}, t, \epsilon, \tau)$, advisor-proposed settings $\bm{\phi}_m$, and read-only \textit{InfoAgent} cues and $\mathcal{H}$ hints.

Each attack method is run with hyperparameters $\bm{\phi}_m$, and enforces box constraints internally. CW and JSMA also enforce the $\ell_\infty$ budget ($\epsilon$) internally. STA relies on its flow smoothness prior ($\mathcal{L}_{\mathrm{flow}}$) for distortion control. \textit{InfoAgent} masks may be used as soft priors but never override $\epsilon$ or $\tau$.

The output $\bm{\delta}_m$ is always projected via $\Pi_{\infty,\epsilon}$ and $\Pi_{[0,1]}$ to ensure compliance with the Phase 1 $\ell_\infty$ budget of $\epsilon$, this includes $\bm{\delta}_{\mathrm{STA}}$, and is accompanied by lightweight diagnostics. Hyperparameters are tuned between but not within runs.

Finally, \textit{MethodAgents} publish compact hints to $\mathcal{H}$ (e.g., active indices), but do not coordinate parameters directly. Their focus is $\bm{\delta}_m$ generation, and mixing is deferred to Phase 4.

\subsection{Phase 3: Critique and Strategic Adaptation}

ARMOR uses a continuous feedback loop. After forming a candidate $\bm{x}_{\text{master}}^{(k)}$ via \eqref{eq:mix}, \textit{CritiqueAgents} evaluate it, appending a feedback vector $\bm{c}^{(k)} \in \mathbb{R}^3$ to history $\mathcal{H}$:
\begin{equation}
\bm{c}^{(k)} = \left[ p_{\text{bb}}(t\,|\,\bm{x}_{\text{master}}^{(k)}), \ \bar{p}_{\text{surr}}(t\,|\,\bm{x}_{\text{master}}^{(k)}), \ \mathrm{SSIM}(\bm{x},\,\bm{x}_{\text{master}}^{(k)}) \right]^{\top}.
\label{eq:critique}
\end{equation}
The components of $\bm{c}^{(k)}$  measure black-box success, transfer stability, and perceptual quality.

\textbf{Local Adaptation.} Each \textit{MethodAgent}'s hyperparameters $\bm{\phi}_m$ (defined in Section~\ref{sec:preliminaries}) are tuned by the \textit{AdvisorAgent}, which reads $\mathcal{H}$ and proposes an updated configuration $\bm{\phi}'_m$. Concretely, $\bm{\phi}_{\mathrm{CW}} = (c,\eta)$ for the CW method, $\bm{\phi}_{\mathrm{JSMA}} = (\alpha,k)$ for JSMA, and $\bm{\phi}_{\mathrm{STA}} = (\gamma, T_{\mathrm{sta}}, S_{\min}, S_{\max}, \theta)$ for STA.

\textbf{Global Strategy \& Escalation.} A central \textit{StrategistAgent} sets high-level policy, mixer weights, and detects stagnation over a sliding window of size $W\in\mathbb{N}$. Let $I_k=\{k-W+1,\ldots,k\}$ denote the window indices, and define:
\begin{equation}
p_{\text{bb}}^{(i)} \triangleq p_{\text{bb}}(t\,|\,\bm{x}_{\text{master}}^{(i)}).
\label{eq:pbb_shorthand}
\end{equation}
Stagnation is detected when both the maximum confidence is below the threshold $\kappa\in(0,1)$ and the improvement range is below $\varrho>0$:
\begin{equation}
\begin{aligned}
&\left(\max_{i\in I_k} p_{\text{bb}}^{(i)}<\kappa\right)\ \land\
 \left(\max_{i\in I_k} p_{\text{bb}}^{(i)}-\min_{i\in I_k} p_{\text{bb}}^{(i)}<\varrho\right) \\
&\Rightarrow\
\begin{cases}
\tau \leftarrow \max(\tau_{\min},\,\tau-\Delta_\tau),\\
\epsilon \leftarrow \min(\epsilon_{\max},\,\epsilon+\Delta_\epsilon),
\end{cases}
\end{aligned}
\label{eq:escalate}
\end{equation}
where $\Delta_\tau,\Delta_\epsilon>0$ are step sizes and $\tau_{\min},\epsilon_{\max}$ are safety bounds. This relaxes the perceptual constraint ($\tau$) and increases the perturbation budget ($\epsilon$), allowing the system to escape local plateaus.

\subsection{Phase 4: Adaptive Perturbation Ensemble}

The final adversarial example $\bm{x}_{\text{master}}^{(k)}$ for the current iteration is synthesized by a \textit{MixerAgent} that adaptively mixes the perturbations given a weight vector $\bm{w}\in\Delta^3$:
\begin{equation}
\bm{\delta}_{\text{mix}}(\bm{w})=\sum_{m=1}^{3} w_m\bm{\delta}_m.
\label{eq:mix_sum}
\end{equation}
We have:
\begin{equation}
\bm{x}_{\text{master}}^{(k)}(\bm{w^*})=\Pi_{[0,1]}\left(\bm{x}+\Pi_{\infty,\epsilon}\big(\bm{\delta}_{\text{mix}}(\bm{w}^{\star})\big)\right),
\label{eq:mix}
\end{equation}
where $\bm{w}^{\star}$ is found by optimizing the score function:
\begin{equation}
\begin{split}
S(\bm{w}) = \quad & \underbrace{\lambda\, p_{\text{bb}}(t \mid \bm{x}_{\text{master}}(\bm{w}))}_{\text{Black-box Success}} + \underbrace{(1-\lambda)\,\bar{p}_{\text{surr}}(t \mid \bm{x}_{\text{master}}(\bm{w}))}_{\text{Transfer Stability}} \\
& - \underbrace{\mu\,\max\big(0,\,\tau-\mathrm{SSIM}(\bm{x},\bm{x}_{\text{master}}(\bm{w}))\big)}_{\text{Quality Penalty}},
\end{split}
\label{eq:score}
\end{equation}
via randomized hill climbing over $\Delta^3$. The score function is an affine combination of the black-box feedback $p_{\text{bb}}(t \mid \cdot)$ and the surrogate ensemble stability $\bar{p}_{\text{surr}}(t \mid \cdot)$. Here, $\lambda\in(0,1)$ is a priority weight and $\mu>0$ is the penalty for violating the SSIM threshold $\tau$.

\renewcommand{\arraystretch}{0.95}
\section{Experimental Evaluation} \label{sec:results}

\begin{table*}[t]
\centering
\caption{Performance of attacks across surrogate and blind detectors, evaluated by Attack Success ASR, wASR, and SSIM, with results reported alongside the standard deviation ($\text{StD}$) for each metric.}
\begin{adjustbox}{max width=\linewidth}
\begin{tabular}{llccccccccc}
\toprule
\multirow{3}{*}{\textbf{Attacker}} &
\multirow{3}{*}{\textbf{Attack Methodology}} &
\multicolumn{6}{c}{\textbf{Surrogate Models}} &
\multicolumn{3}{c}{\textbf{Target Blind Model}} \\
\cmidrule(lr){3-8}\cmidrule(lr){9-11}
 & & \multicolumn{3}{c}{ResNet-50} & \multicolumn{3}{c}{DenseNet-121} & \multicolumn{3}{c}{ViT-B/16} \\
\cmidrule(lr){3-5}\cmidrule(lr){6-8}\cmidrule(lr){9-11}
 & & \textbf{ASR $\pm\text{StD}$} & \textbf{wASR $\pm\text{StD}$} & \textbf{SSIM $\pm\text{StD}$} &
     \textbf{ASR $\pm\text{StD}$} & \textbf{wASR $\pm\text{StD}$} & \textbf{SSIM $\pm\text{StD}$} &
     \textbf{ASR $\pm\text{StD}$} & \textbf{wASR $\pm\text{StD}$} & \textbf{SSIM $\pm\text{StD}$} \\
\midrule
\rowcolor{gray!15}\multicolumn{11}{l}{\textit{Transfer-based black-box attacks}} \\
MI-FGSM~\cite{8579055}  & Momentum iterative           
& 0.994$\pm$0.075          & 0.743$\pm$0.065 & 0.747$\pm$0.034
& \textbf{1.000$\pm$0.000} & 0.747$\pm$0.034 & 0.747$\pm$0.034
& 0.068$\pm$0.251          & 0.051$\pm$0.189 & 0.747$\pm$0.034 \\
DI-FGSM~\cite{xie2019improving} & Diverse input                
& 0.906$\pm$0.293          & 0.768$\pm$0.250 & 0.848$\pm$0.031
& 0.997$\pm$0.053          & 0.845$\pm$0.055 & 0.848$\pm$0.031
& 0.038$\pm$0.191          & 0.032$\pm$0.163 & 0.848$\pm$0.031 \\
TI-FGSM~\cite{dong2019evading}  & Translation invariant        
& 0.713$\pm$0.453          & 0.648$\pm$0.413 & 0.907$\pm$0.034
& 0.903$\pm$0.296          & 0.821$\pm$0.271 & 0.907$\pm$0.034
& 0.151$\pm$0.358          & 0.136$\pm$0.324 & 0.907$\pm$0.034 \\
SINI-FGSM~\cite{Lin2020Nesterov} & Sinusoidal noise             
& 0.935$\pm$0.246          & 0.668$\pm$0.179 & 0.714$\pm$0.037
& \textbf{1.000$\pm$0.000} & 0.714$\pm$0.037 & 0.714$\pm$0.037
& 0.059$\pm$0.236          & 0.042$\pm$0.169 & 0.714$\pm$0.037 \\
\midrule
\rowcolor{gray!15} \multicolumn{11}{l}{\textit{Query-based black-box attacks}} \\
Square~\cite{andriushchenko2020square} & Random search                
& 0.000$\pm$0.000          & 0.000$\pm$0.000 & 0.884$\pm$0.082
& 0.023$\pm$0.149          & 0.020$\pm$0.133 & 0.884$\pm$0.082
& 0.008$\pm$0.092          & 0.008$\pm$0.083 & 0.884$\pm$0.082 \\
SimBA-DCT~\cite{guo2019simple}  & Coordinate / frequency query 
& 0.154$\pm$0.361          & 0.147$\pm$0.345 & 0.970$\pm$0.030
& 0.176$\pm$0.381          & 0.168$\pm$0.365 & 0.970$\pm$0.030
& 0.004$\pm$0.065          & 0.004$\pm$0.061 & 0.970$\pm$0.030 \\
PBO~\cite{10.5555/3692070.3692392}  & Bayesian optimization (function prior) 
& 0.000$\pm$0.000          & 0.000$\pm$0.000 & 0.972$\pm$0.029
& 0.000$\pm$0.000          & 0.000$\pm$0.000 & 0.972$\pm$0.029
& 0.003$\pm$0.053          & 0.003$\pm$0.050 & 0.972$\pm$0.029 \\
\midrule
\rowcolor{gray!15} \multicolumn{11}{l}{\textit{Ensemble-based black-box attacks}} \\
AutoAttack-PGD~\cite{croce2020reliable}    & Projected gradient ensemble        
& \textbf{1.000$\pm$0.000} & 0.552$\pm$0.035 & 0.552$\pm$0.035
& \textbf{1.000$\pm$0.000} & 0.573$\pm$0.037 & 0.573$\pm$0.037
& 0.196$\pm$0.397          & 0.108$\pm$0.220 & 0.563$\pm$0.037 \\
AutoAttack-Square~\cite{croce2020reliable} & Query-based ensemble              
& 0.976$\pm$0.153          & 0.708$\pm$0.178 & 0.728$\pm$0.141
& 0.959$\pm$0.198          & 0.684$\pm$0.194 & 0.719$\pm$0.136
& 0.076$\pm$0.265          & 0.047$\pm$0.169 & 0.719$\pm$0.141 \\
MSGAGA~\cite{msgaga2025}        & Metric-Guided Selection            
& \textbf{1.000$\pm$0.000} & 0.760$\pm$0.129 & 0.760$\pm$0.129
& \textbf{1.000$\pm$0.000} & 0.760$\pm$0.129 & 0.760$\pm$0.129
& 0.061$\pm$0.239          & 0.052$\pm$0.205 & 0.760$\pm$0.129 \\
\midrule
\rowcolor{gray!15} \multicolumn{11}{l}{\textit{Agentic-based black-box attacks}} \\
RL-PPO~\cite{domico2025adversarial} & Reinforcement-learning agent           
& 0.000$\pm$0.000          & 0.000$\pm$0.000          & 0.973$\pm$0.029
& 0.000$\pm$0.000          & 0.000$\pm$0.000          & 0.973$\pm$0.029
& 0.003$\pm$0.053          & 0.003$\pm$0.050          & \textbf{0.973$\pm$0.029} \\
\textbf{ARMOR (ours)} & Multi-Agent Orchestration      
& \textbf{1.000$\pm$0.000} & \textbf{0.982$\pm$0.032} & \textbf{0.982$\pm$0.032}
& \textbf{1.000$\pm$0.000} & \textbf{0.977$\pm$0.039} & \textbf{0.977$\pm$0.039}
& \textbf{0.396$\pm$0.489} & \textbf{0.280$\pm$0.362} & 0.701$\pm$0.170 \\
\bottomrule
\end{tabular}
\label{tab:tab_1}
\end{adjustbox}
\end{table*}

\begin{figure*}[t]
    \centering
    \includegraphics[width=\linewidth]{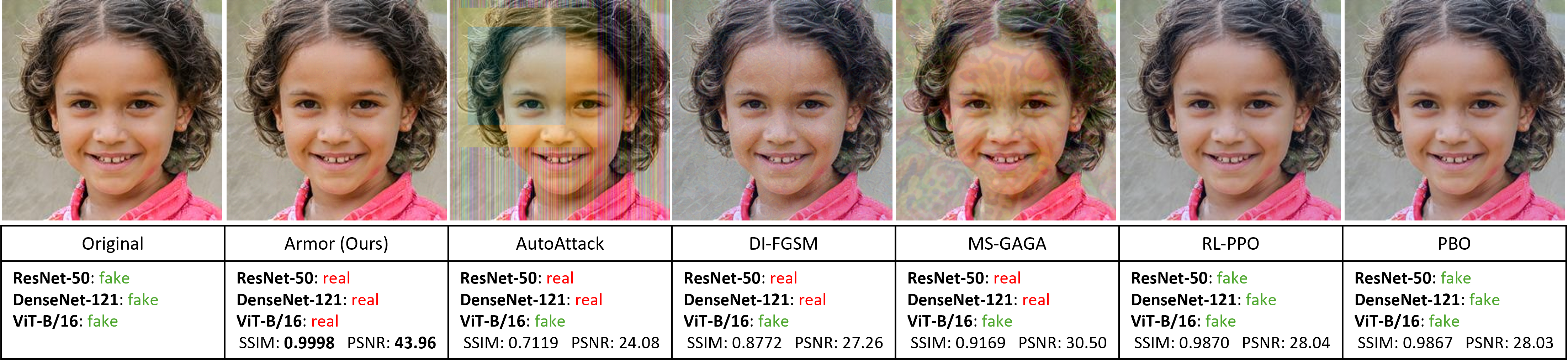}
    \caption{Comparison of adversarial examples generated by different methods on a representative fake image from the AADD-LQ dataset. Green and red labels indicate correct and incorrect predictions, respectively, across three detectors.
    }
    \label{fig:armor}
\end{figure*} 

\renewcommand{\arraystretch}{0.99}
\begin{table}[t]
\centering
\caption{Transferability of attacks with surrogate success rate $p_\text{surr} > 0.5$, evaluated using metrics defined in Eq.~(\ref{eq:psurr})--(\ref{eq:pcond}) and $\Delta \text{ASR}$, which represents the drop in attack success rate from surrogate to blind model.
}
\begin{adjustbox}{max width=\linewidth}
\begin{tabular}{lcccc}
\toprule
\textbf{Attacker (with $p_\text{surr}>0.5$)} &
$\bm{p_{\text{surr}}}$ Eq.~(\ref{eq:psurr}) $\uparrow$ &
$\bm{p_{\text{tgt}}}$ Eq.~(\ref{eq:ptgt}) $\uparrow$ &
$\bm{p_{\text{cond}}}$ Eq.~(\ref{eq:pcond}) $\uparrow$ &
$\bm{\Delta\text{ASR}}$  $\downarrow$ \\
\midrule
TI-FGSM~\cite{dong2019evading}   & 0.914 & 0.151 & 0.165 & 0.657 \\
MI-FGSM~\cite{8579055}           & 1.000 & 0.068 & 0.068 & 0.930 \\
DI-FGSM~\cite{xie2019improving}  & 0.999 & 0.038 & 0.038 & 0.913 \\
SINI-FGSM~\cite{Lin2020Nesterov} & 1.000 & 0.059 & 0.059 & 0.908 \\
\midrule
AutoAttack-PGD~\cite{croce2020reliable}    & 1.000 & 0.196 & 0.196 & 0.804 \\
AutoAttack-Square~\cite{croce2020reliable} & 0.999 & 0.076 & 0.076 & 0.892 \\
MSGAGA~\cite{msgaga2025}                  & 1.000 & 0.061 & 0.061 & 0.939 \\
\midrule
\textbf{ARMOR (ours)} & \textbf{1.000} & \textbf{0.396} & \textbf{0.396} & \textbf{0.604} \\
\bottomrule
\end{tabular}
\label{tab:tab_2}
\end{adjustbox}
\end{table}

\begin{table}[t]
\centering
\caption{Ablation study of the ARMOR framework on the blind model.
Each configuration is evaluated on AADD-LQ using ASR (\%), wASR, and SSIM.}
\begin{adjustbox}{max width=\linewidth}
\begin{tabular}{lccc}
\toprule
\textbf{Configuration} & \textbf{ASR} & \textbf{wASR} & \textbf{SSIM} \\
\midrule
\textbf{ARMOR (Full)} &
\textbf{0.396$\pm$0.489} & \textbf{0.280$\pm$0.362} & \textbf{0.701$\pm$0.170} \\
\midrule
\textit{(a) Uniform Averaging (No Orchestration)} &
0.010$\pm$0.099 & 0.010$\pm$0.095 & 0.959$\pm$0.014 \\
\textit{(b) Without Info Agent (No Reasoning)} &
0.006$\pm$0.075 & 0.005$\pm$0.061 & 0.810$\pm$0.027 \\
\textit{(c) Without Info + Conductor (No Reparam.)} &
0.007$\pm$0.084 & 0.006$\pm$0.068 & 0.803$\pm$0.029 \\
\bottomrule
\end{tabular}
\end{adjustbox}
\label{tab:abla}
\end{table}

\subsection{Setup}
We evaluate ARMOR on the low-quality (LQ) fake subset of the AADD benchmark~\cite{bongini2025wild}, which contains 710 fake images from diverse content synthesis engines. All images are labeled as fake, resized to $224 \times 224$, and normalized using standard ImageNet statistics (mean = [0.485, 0.456, 0.406], std = [0.229, 0.224, 0.225]).

\textbf{Attack Baselines.}  
We benchmark ARMOR against various adversarial attacks grouped by their attack style: (i) transfer-based (MI-FGSM~\cite{8579055}, TI-FGSM~\cite{dong2019evading}, DI-FGSM~\cite{xie2019improving}, SINI-FGSM~\cite{Lin2020Nesterov}); (ii) query-based (SimBA-DCT~\cite{guo2019simple}, Square~\cite{andriushchenko2020square}, PBO~\cite{10.5555/3692070.3692392}); (iii) ensemble-based (AutoAttack variants~\cite{croce2020reliable}, PGD~\cite{madry2018towards}, Square~\cite{andriushchenko2020square}, MSGAGA~\cite{msgaga2025}); and (iv) agentic-based (L-PPO~\cite{domico2025adversarial}).
All white-box variants craft perturbations on surrogate models (ResNet-50 / DenseNet-121) and are then evaluated on the blind target (ViT-B/16). For fair comparison, we use a common
$\ell_\infty$ budget of $\epsilon=8/255$ and identical query across methods.

\textbf{Evaluation Metrics.}
The Attack Success Rate (ASR), Weighted ASR (wASR), and SSIM are reported.
ASR measures the proportion of adversarial samples that fool a detector. 
Let $\mathbbm{1}[\cdot]$ denotes the indicator function, then $\text{ASR} = \frac{1}{N} \sum_{i=1}^{N} \mathbbm{1}[\hat{y}_i \neq y_i]$.
For imperceptibility, wASR reweights each successful attack by its local SSIM, $\text{wASR} = \frac{1}{N} \sum_{i=1}^{N} \text{SSIM}_i \cdot \mathbbm{1}[\hat{y}_i \neq y_i]$,
with SSIM quantifying perceptual similarity.

To assess transferability, we compute: surrogate success rate $p_{\text{surr}}$, target success rate $p_{\text{tgt}}$, and conditional transfer probability $p_{\text{cond}}$.
Let $\hat{y}i^{m}$ be the prediction of model $m$ on image $i$ and $y_i$ the ground truth.
An attack succeeds when $\hat{y}i^{m} \neq y_i$:
\begin{equation}
p_{\text{surr}} = \frac{1}{N} \sum{i=1}^{N}
\mathbbm{1}\big[(\hat{y}i^{RN50} \neq y_i) \lor (\hat{y}i^{DN121} \neq y_i)\big],
\label{eq:psurr}
\end{equation}

\begin{equation}
p_{\text{tgt}} = \frac{1}{N} \sum{i=1}^{N}
\mathbbm{1}[\hat{y}i^{ViT} \neq y_i],
\label{eq:ptgt}
\end{equation}

\begin{equation}
p_{\text{cond}} =
\frac{
\mathbb{E}\left[\mathbbm{1}\big[(\hat{y}_i^{ViT} \neq y_i) \land ((\hat{y}_i^{RN50} \neq y_i) \lor (\hat{y}_i^{DN121} \neq y_i))\big]\right]
}{
\mathbb{E}\left[\mathbbm{1}\big[(\hat{y}_i^{RN50} \neq y_i) \lor (\hat{y}_i^{DN121} \neq y_i)\big]\right]
}.
\label{eq:pcond}
\end{equation}
Here, $p_{\text{surr}}$ measures success on at least one surrogate, $p_{\text{tgt}}$ on the blind model, and $p_{\text{cond}}$ quantifies the likelihood that an attack successful on a surrogate also transfers to the target.

\textbf{Computation Resources.} 
Experiments are conducted on a workstation with NVIDIA RTX~4090 GPUs and AWS EC2 g6e.8xlarge instances for the \textit{InfoAgent} and \textit{ConductorAgent}. Metrics are averaged over three random seeds. Hyperparameters: CW iterations = 1000, JSMA top-$k$ = 2, STA regularization = 0.1, Mixer hill-climb iterations = 500, max queries per image = 10k.

\subsection{Results}
We evaluate each attack along two dimensions:
(i) attack success vs. perceptual quality, and
(ii) surrogate success vs. transferability.

\textbf{Attack Success vs. Perceptual Quality.}  
As shown in Table~\ref{tab:tab_1}, ARMOR achieves the highest attack success across both surrogate and blind models. On surrogate detectors, it attains a perfect ASR of 1.000 with high wASR (0.982, 0.977), indicating complete success with minimal perceptual degradation. When transferred to the blind ViT model, ARMOR maintains strong ASR (0.396) and wASR (0.280) but slightly lower SSIM (0.701), reflecting a deliberate trade-off favoring higher attack success over visual fidelity. In contrast, transfer-based methods (MI-FGSM, DI-FGSM, TI-FGSM) show sharp performance drops on the blind model, revealing poor transferability, while query- and ensemble-based attacks (Square, SimBA-DCT, AutoAttack) achieve high SSIM yet fail to deceive effectively. Due to space constraints, Fig.~\ref{fig:armor} presents representative examples, where ARMOR’s perturbations remain nearly imperceptible yet consistently flip all detector outputs from “fake” to “real”. 

\textbf{Surrogate Success vs. Transferability.}
The transferability analysis in Table~\ref{tab:tab_2} shows that traditional transfer-based attacks such as TI-FGSM, DI-FGSM, and MI-FGSM exhibit high surrogate success ($p_\text{surr} \approx 1.0$) but transfer poorly to the blind ViT model, with low conditional transfer probabilities ($p_\text{cond} < 0.2$). Ensemble-based approaches (e.g., AutoAttack-PGD, AutoAttack-Square) improve slightly in $p_\text{tgt}$ yet still suffer from large $\Delta\text{ASR}$ gaps. In contrast, ARMOR attains perfect surrogate success ($p_\text{surr} = 1.0$) and the highest conditional transfer probability ($p_\text{cond} = 0.396$), over twice that of the next best method. Although the transfer drop ($\Delta\text{ASR}=0.604$) reflects the difficulty of generalizing to unseen architectures, ARMOR’s agentic orchestration still yields the strongest transfer, supporting its robustness as a black-box attack framework.

\textbf{Ablation Study}
Table~\ref{tab:abla} measures each ARMOR component on the blind ViT model. The full system attains the highest ASR and wASR, confirming that multi-agent orchestration together with dynamic reparameterization yields effective, transferable attacks. Replacing orchestration with averaging (case a) causes a drop in ASR/wASR, showing that adaptive coordination among \emph{MethodAgents} is essential.
Disabling the \emph{InfoAgent} (case b) further reduces success, as the \emph{ConductorAgent} lacks contextual reasoning about image artifacts to adjust its strategy.
Removing both Info and Conductor (c) slightly outperforms (b) by reverting to stable default settings. Overall: (i) orchestration is essential, (ii) reasoning must be coupled with reliable reparameterization, and (iii) unguided adaptation can reduce transfer.

\section{Conclusions} \label{sec:conclusions}

\noindent ARMOR has introduced a VLM-guided multi-agent framework that has orchestrated CW, JSMA, and STA attacks with closed-loop critique and adaptive mixing. It has achieved perfect surrogate success and stronger transfer to a blind ViT-B/16 while keeping perceptual quality competitive. Ablations have demonstrated that orchestration and reparameterization are essential. Future work will consider improving perceptual quality, broadening attack families, and defenses that use cross-agent signals.

\bibliographystyle{IEEEbib}
\bibliography{refs}

\end{document}